\def\year{2021}\relax
\documentclass[letterpaper]{article} 
\usepackage{aaai21}  
\usepackage{times}  
\usepackage{helvet} 
\usepackage{courier}  
\usepackage[hyphens]{url}  
\usepackage{graphicx} 
\def\UrlFont{\rm}  
\usepackage{natbib}  
\usepackage{caption} 
\nocopyright
\usepackage{subcaption}

\usepackage{todonotes}
\usepackage{amsmath}
\usepackage{amsthm}
\usepackage{amsfonts}
\DeclareMathOperator*{\argmax}{arg\,max}
\DeclareMathOperator*{\argmin}{arg\,min}
\usepackage{algorithm}
\usepackage{algorithmic}
\usepackage[switch]{lineno}

\title{Divide and Learn: A Divide and Conquer Approach for Predict+Optimize}
\author {
    Authors

        Ali Ugur Guler,\textsuperscript{\rm 1}
        Emir Demirovic, \textsuperscript{\rm 2}
        Jeffrey Chan, \textsuperscript{\rm 3}
        James Bailey, \textsuperscript{\rm 1}
        Christopher Leckie, \textsuperscript{\rm 1}
        Peter J. Stuckey, \textsuperscript{\rm 4}\\
}
\affiliations {
    \textsuperscript{\rm 1} University of Melbourne, 
    \textsuperscript{\rm 2} Delft University of Technology, 
    \textsuperscript{\rm 3} RMIT University, 
    \textsuperscript{\rm 4} Monash University
    aguler@student.unimelb.edu.au, e.demirovic@tudelft.nl, jeffrey.chan@rmit.edu.au, baileyj@unimelb.edu.au, caleckie@unimelb.edu.au, peter.stuckey@monash.edu.au  
}
\begin{document}
\maketitle
\newcommand{\jb}[1]{\textcolor{blue}{JB: #1}}
\newcommand{\aug}[1]{\textcolor{purple}{AUG: #1}}
\newcommand{\augedit}[1]{\textcolor{purple}{#1}}
\newcommand{\jc}[1]{\textcolor{green}{JC: #1}}
\newcommand{\pjs}[1]{\textcolor{red}{PJS: #1}}

\newcommand{\ignore}[1]{}

\newcommand{\vopt}[0]{\pmb{v}}
\newcommand{\voptsmall}[0]{v}
\newcommand{\voptbig}[0]{V}

\newcommand{\vpred}[0]{\pmb{v_p{}}}
\newcommand{\vpredsmall}[0]{v_p{}}

\newcommand{\xopt}[0]{\pmb{x}}
\newcommand{\xoptsmall}[0]{x}
\newcommand{\xpred}[0]{\pmb{x_p{}}}
\newcommand{\xpredsmall}[0]{x}

\newcommand{\vparam}[0]{\pmb{\theta}}
\newcommand{\param}[0]{\theta}
\newcommand{\coef}[0]{\beta}

\newcommand{\vcoef}[0]{\pmb{\beta}}

\newtheorem{theorem}{Theorem}
\newtheorem{corollary}{Corollary}
\newtheorem{lemma}[theorem]{Lemma}
\newtheorem*{proposition}{Proposition}

\theoremstyle{definition}
\newtheorem{exmp}{Example}

\begin{abstract}
The predict+optimize problem combines machine learning of problem coefficients with a
combinatorial optimization problem that uses the predicted coefficients. 
While this problem can be solved in two separate stages, 
it is better to directly minimize the optimization loss. 
However, this requires differentiating through a discrete,
non-differentiable combinatorial function.
Most existing approaches use some form of surrogate gradient. 
Demirovic \emph{et al} showed how to directly express the loss of the optimization
problem in terms of the predicted coefficients as a
piece-wise linear function.  However, their approach is restricted
to optimization problems with a dynamic
programming formulation. 
In this work we propose a novel
divide and conquer algorithm to tackle optimization problems without this restriction 
and predict
its coefficients using the optimization loss.
We also introduce a greedy version of this approach, which achieves similar
results with less computation. 
We compare our approach with other approaches to the predict+optimize
problem and show we can successfully tackle some hard combinatorial problems
better than other predict+optimize methods.
\ignore{
Combinatorial optimization is key for successful decision making. In many real-world applications, parameters for the optimization problem are not known and have to be predicted. Traditionally a two-staged framework is used to predict and then optimize. The first stage, a Machine Learning(ML) model, predicts the parameters using a conventional loss, such as the mean squared error(MSE).
The second stage, a solver, finds a solution to the predicted optimization problem. This creates a discrepancy between the training loss and the test loss. Predict+optimize aims to train models to minimize optimization loss. However, this requires differentiating through a discrete, non-differentiable function. A popular approach is to use a differentiable surrogate function and train the model with a surrogate gradient \cite{elmachtoub2017smart}\cite{wilder2019melding}\cite{mandi2019smart}. Although surrogates can capture easy parts of the optimization problems, they might fail to understand harder problems. \cite{demirovicdynamic} constructs a direct representation of the optimization problem as a piece-wise linear(PWL) function. But their approach relies on dynamic programming to map the PWL functions and therefore it's application areas are limited to a small number of optimization problems. We propose a novel divide and conquer algorithm to map the optimization problem and predict it's parameters using the optimization loss. Our model can directly reason over the optimization problem and can generalize into problems without a DP solution. We also introduce greedy algorithms that partially map the problem and still achieves similar performance to complete methods. We experiment on knapsack and energy cost-aware scheduling problems and compare our model with the state of the art SPO-relax model. We show that our model can successfully capture hard combinatorial problems where SPO-relax can fail to do so. 
}
\end{abstract}

\section{Introduction}

Machine Learning (\textbf{ML}) has gained substantial attention in the last decade, and has proven to be useful in a wide range of industries. ML models usually focus on making accurate predictions by minimizing errors, such as mean squared error (\textbf{MSE}). These predictions can then be used as coefficients in other decision making processes, such as a combinatorial optimization problem. The real performance of these predictions is evaluated by their ability to lead to the correct decisions. Such evidence based decision making arises in many fields like transportation, healthcare, security and education \cite{horvitz2010data}. Consequently, there has been growing interest in ML models for use in optimization problems. These models try to predict coefficients of the optimization problem in such a way that even if the predictions are less accurate, they lead to better decisions. This paradigm is called \textbf{predict then optimize} \cite{elmachtoub2017smart} or \textbf{predict+optimize} \cite{aaai20e}. In this paper we propose a new framework for predict+optimize to learn coefficients by directly reasoning over discrete combinatorial problems.

\textbf{Motivation:} Traditionally, ML models treat predictions as the end goal. For example, a regression model will try to minimize the MSE of its predictions. However, if these predictions are the coefficients of an optimization problem, the prediction and optimization tasks are not independent operations. 

\begin{exmp}\label{ex_candidates}
\emph{There are two open research positions at a prestigious institute, and three candidates for the two positions. The principal investigator (PI) needs to decide which two candidates to choose. In order to make this decision, the PI needs to predict how productive each candidate will be, given information about the candidates, like their past papers and institutions. Suppose the PI designs two models to predict how many papers each candidate is likely to publish every year. The first model is a traditional ML model. It learns to minimize the MSE of the predicted number of publications for each candidate. The second model is a predict+optimize model that learns how to pick the most productive two candidates. Suppose these three candidates $a$,$b$,$c$ will publish $\voptsmall_a=3$, $\voptsmall_b=5$, $\voptsmall_c=7$ papers a year, respectively. The ML model predicts that the candidates will publish $\vpredsmall_a=5$, $\vpredsmall_b=4$, $\vpredsmall_c=6$ papers a year. In contrast, the predict+optimize model seems to make inaccurate predictions about the candidates, with $\vpredsmall_a=0$, $\vpredsmall_b=50$, $\vpredsmall_c=70$. The PI thinks predict+optimize predictions are unlikely and they decide to use the ML model' s predictions.}
\end{exmp}

In this example the ML model makes noticeably more accurate predictions compared to the predict+optimize model. If we evaluate their performance with regards to MSE, then the ML model outperforms the 
predict+optimize model by an MSE of 2 to 2001. In contrast, if we evaluate the models' performance using the result of the optimization problem, predict+optimize outperforms ML by choosing the more productive candidates $\{b,c\}$ over the less productive candidates $\{a,c\}$. By choosing a standard ML model, the PI fails to realize that predict+optimize reasoned over the optimization problem, and learned the underlying ranking of the candidates. The predict+optimize model penalized the least productive candidate, and exaggerated the most productive candidates. In fact these ``inaccurate'' predictions were helpful in making the correct decision. 

If the ML model makes perfect predictions, it also leads to the optimal decision. However all models are prone to errors. When there are errors in predictions, MSE does not necessarily represent the performance of the decisions \cite{ifrim2012properties}. For this candidate selection problem, penalizing errors that change the relative productivity is more important and the ML model failed to penalize errors that disturbs the relative ordering of the candidates. A predict+optimize framework trains parameters with respect to the optimization objective, and it can understand the underlying problem better. 
Although  predict+optimize  can improve  decision  making,  these  models  require  learning through  hard,  often  non differentiable  and  discrete  functions.

\ignore{
\citeauthor{ijcai19d}~(\citeyear{ijcai19d}) categorizes three approaches to tackle the predict+optimize challenge: indirect, semi-direct and direct approaches. 

\textbf{Indirect approaches} consider prediction and optimization as separate entities. During the prediction stage a ML model predicts optimization coefficients independently from the optimization problem. The main goal of these approaches is to increase individual predictions, and with accurate enough predictions, they will lead to correct decisions. However if the difference between the underlying optimization problem and MSE is large, indirect approaches may fail to capture this problem. Different ML models can be tried to choose the one with the best optimization performance.

\textbf{Semi-direct approaches} combine heuristics with indirect approaches. Similar to the indirect approaches the ML model consider the optimization and during prediction as separate. However if an expert knows the underlying optimization problem, they can modify the ML model to predict coefficients for that specific problem. For example in the candidate selection problem, if the PI knows that the underlying problem is a ranking problem, they can choose to train a model to rank the candidates. These solutions are often very specialized and are only applicable to very specific optimization problems. When the optimization problems get complex, small differences in constraints can change the solution space dramatically. Therefore for each different set of constraints, and problem sets different heuristics have to be prepared individually. For well known problems it might easier to develop these heuristic, for the niche problems the cost to develop these heuristics can be high.


\textbf{Direct approaches} train their model parameters directly on the optimization loss. The aim is to build a framework that understands the underlying problem for an arbitrary optimization problem without heuristics. A lot of decision making processes rely on combinatorial optimization. In such cases a direct predict+optimize model needs to differentiate through a non-differentiable, discrete optimization problem. 
}
One way to differentiate through combinatorial problems is to use surrogates, however surrogates induce an approximation error to the optimization objective~\cite{thapper2018limits}. \citeauthor{aaai20e}~(\citeyear{aaai20e}) propose a novel framework to directly reason over the \emph{exact} optimization loss for 
problems with a linear objective 
with a dynamic programming (DP) solution. They
represent the optimization coefficients as parameterised linear functions and use the parameterised linear functions to solve the optimization problem with DP. The DP solver performs piece-wise linear algebra to construct a piece-wise linear function (PWLF). They show that the transition points of the PWLF can capture the underlying optimization loss. Their model uses the transition points to train model parameters, and they achieve improvements for knapsack problems. Although this approach can understand the exact optimization problem, it relies on DP, and therefore it's application areas are limited only to problems with a DP solution. Their frameworks exhaustively finds all the transition points. However, large and complex problems may have a large number of 
uninteresting transition points. For such problems the DP approach may fail to scale well, and take too long to run.

In this paper we propose a novel framework to directly reason over the exact optimization loss (with no restriction to DP). Our framework builds upon the idea of representing the optimization loss as a PWLF. However, unlike the DP solution, we use a numerical approach to extract transition points. We show that the predicted PWLF is a convex function. 
We use this knowledge to apply a divide and conquer algorithm to compare different sample points and identify transition points. We then evaluate the transition points on the real optimization problem and train our model parameters with the exact optimization loss. We further propose greedy methods and show that less accurate transition point identification can decrease the run time, and still achieve similar performance to the full methods.
First we experiment on 0-1 knapsack problems with both unit weights and varying weights. We show that our divide and conquer approach achieves identical results to the DP model and scales better for larger problems. We also demonstrate that for hard knapsack problems the exact methods are more robust compared to state of the art surrogate models (SPO-Relax, QPTL). To demonstrate our framework's ability to reason with arbitrary optimization problems, we experiment on a complex scheduling problem, which the DP method is not able to reason about. 

Our contributions are as follows:
\begin{itemize}
    \item Show for optimization problems with linearly parameterised coefficients and linear objectives, predicted objective value is convex.
    \item A novel framework to directly reason over the exact optimization problem based on PWLF mapping. Unlike the previous state of the art DP method, our framework is not limited to DP and can be used for any arbitrary optimization problems with a linear objective.
    \item Greedy methods that show that a less than perfect PWLF mapping still achieve similar performance to a full mapping, and reduce the run time considerably. We show that our greedy methods scales better than the previous DP approach for larger problems.
    \item Evaluation on 0-1 knapsack problem and a non DP scheduling problem, and comparison with the previous state of the art exact method DP, and two state of the art surrogate methods QPTL, SPO-Relax.

\end{itemize}

\section{Related Work}
\newcommand{\citeay}[1]{\citeauthor{#1}~(\citeyear{#1})}

The standard approach to predict+optimize problems is to separately solve the prediction problem and then the optimization problem. 
Combined approaches are a relatively new focus. 
\citeauthor{bengio1997using}~(\citeyear{bengio1997using}) showed that for hand crafted models to optimize a financial portfolio, profit performs better than a standard loss function.
\citeauthor{kao2009directed}
~(\citeyear{kao2009directed}) proposed using a combination of Empirical Optimization and ordinary least squares loss to improve performance for decision driven machine learning. 
\citeay{lim2012robust} define relative regret in the context of portfolio optimization.
\citeauthor{elmachtoub2017smart}~(\citeyear{elmachtoub2017smart}) 
define the general Smart Predict and Optimize (SPO) loss, which we call the regret in our paper. They propose a linear relaxation SPO+ loss to train machine learning models. Their work shows SPO+ loss can be used to achieve improved performance for constrained linear programming problems. 
\citeay{amos2017optnet} propose to transform the optimization loss into a quadratic problem using KKT equations. \citeay{donti2017task} show that performance can be improved by using sequential quadratic programming (QP) to compute the new loss, and train the model with respect to it. 
\citeay{wilder2019melding} extend the QP approach to linear programming problems. 
\citeay{ferber2020mipaal} extend the approach of \citeay{wilder2019melding}
to directly apply to mixed integer programming by using pure cutting plane methods to solve the MIP, resulting in an LP sufficient to define the MIP optimally.
\citeay{aaai20a} also show that SPO+ loss can be used as a surrogate loss for relaxations of combinatorial problems and achieve performance improvements.
\citeay{ijcai2020-208} propose a specialised framework to optimize virtual machine provisioning.
Black-box end to end frameworks are also used to differentiate and learn combinatorial problems~\cite{bello2016neural},~\cite{li2018combinatorial},\cite{niculae2018sparsemap}. \citeay{iclr2020Differentiation} use a black-box approach to predict optimal solutions from coefficient features.
\citeay{ijcai19d} investigate the knapsack problem from a predict+optimize perspective  and show how ranking methods can be applied to it. Similarly \citeay{demirovic2019predict+} introduce transition points for ranking problems. 
The direct inspiration of our work is that of \citeay{aaai20e}, which shows how to optimize parameters in a learning model directly using regret, as long as the optimization problem has a dynamic programming formulation.
They build a piecewise linear function using the dynamic programming formulation that identifies transition points, where the regret changes.
In this work we extend this approach to arbitrary optimization problems by using numerical methods to find transition points in the regret loss function.

\section{Divide and Learn}\label{framework}

\ignore{
 We first show that for linear models, predicted objective value is convex. Later we detail transition point extraction and parameter update. Finally we define two greedy versions of our framework, \emph{DnL-MAX} and \emph{DnL-Greedy}.
}

\subsection{Preliminaries}\label{prelim} 
Our framework \emph{divide and learn (DnL)} predicts coefficients with a linear model. We show for linear models, predicted objective value is a convex piecewise function.
We now formally define the predict+optimize problem.
Given a set of objective coefficients
$\vopt$, we define an optimization problem and its solution as:
\begin{equation}
    \max_{\xopt \in C} Obj(\xopt,\vopt)
\end{equation}
\begin{equation}
    s(\vopt)\equiv\argmax_{\xopt\in C} Obj(\xopt,\vopt)
\end{equation}
where $C$ is the set of feasible solutions (usually described implicitly). 
The oracle $s(\vopt)$ finds a solution that
maximizes the objective value of the optimization problem given objective coefficients
$\vopt$.
In predict+optimize settings objective coefficients
are not known beforehand and they are
predicted using features
$\vparam$ and parameters
$\vcoef$, $\vpred = \vpred(\vparam,\vcoef)$.
We show the new parameterized optimization problem as
\begin{equation}\label{eq_oracle}  
     s(\vparam,\vcoef) \equiv \argmax_{\xopt \in C} Obj(\xopt,\vpred(\vparam,\vcoef))
\end{equation}

\textbf{Regret: }We measure the performance of predict+optimize frameworks using regret. Regret is defined as the cost of making sub-optimal decisions due to incorrect coefficient predictions. If we define $\xopt^* = s(\vopt)$ as the optimal solution of an optimization problem with \emph{true} objective coefficients $\vopt$, and $\xpred = s(\vpred)$ as the optimal solution of an optimization problem with predicted objective coefficients $\vpred$, then regret $R$ is:
\begin{equation}\label{eq_reg}
    R(\vpred,\vopt)= Obj(\xopt^*,\vopt) - Obj(\xpred,\vopt)
\end{equation}
The true optimal value $Obj(\xopt^*,\vopt)$ represents a boundary for the best decisions made with the predicted coefficients. The optimal objective value with predicted coefficients $Obj(\xpred,\vopt)$ can never exceed the true optimal. Therefore the minimum value of regret is zero, and it is achieved when the predicted optimal solution, $\xpred$, is equal to the true optimal solution, $\xopt^*$.
The \textbf{predict+optimize problem} is to find $\vcoef$ that minimizes
$R(\vpred(\vparam,\vcoef),\vopt)$.

\textbf{Transition points:} Note that parameterised regret is a piece wise function. The predicted coefficients $\vpred(\vparam,\vcoef)$ can only affect regret by changing the solution of the optimization problem. These changes are not continuous and 
only happen at specific boundaries of the $\vcoef$ values. 
Assume for the moment a single (unfixed) parameter $\beta$.
We call parameter values $\coef_t$ where the optimal solution changes as the \textbf{transition points} of the piece wise regret function. Note that for any two points between consecutive transition points $\coef_{t_{i}}<\coef_1<\coef_2<\coef_{t_{i+1}}$, $s(\vparam,\coef_1) = s(\vparam,\coef_2)$, therefore $ regret(\coef_1)=regret(\coef_2)$. This suggests mapping the optimization problem by identifying intervals defined by the transition points. Then we can choose any value in those intervals to train model parameters.

\begin{figure}
    \caption{Piece wise function construction}
    \centering
    \begin{subfigure}[b]{.13\textwidth}
        \centering
         \includegraphics[width=1\textwidth]{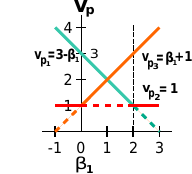}%
        \caption{Predicted coefficient($\vpred$)}
        \label{fig:PWL}
    \end{subfigure}%
     \hspace{1em}
    \begin{subfigure}[b]{.13\textwidth}
        \centering
        \includegraphics[width=0.8\textwidth]{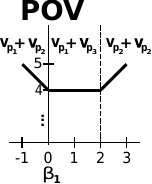}
        \caption{Predicted optimal value(POV)}
        \label{fig:PWLsum}
    \end{subfigure}
    \begin{subfigure}[b]{.13\textwidth}
        \centering
        \includegraphics[width=0.8\textwidth]{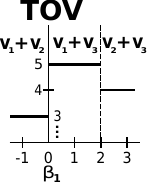}
        \caption{True optimal value(TOV)}
        \label{fig:PWLopt}
    \end{subfigure}
    \label{fig:PWF_construction}
\end{figure}

\begin{exmp}\label{ex_knap}
\emph{Consider a knapsack problem with three items valued at $\voptsmall_{1}=2$, $\voptsmall_{2} = 1$, $\voptsmall_{3} = 3$ . The capacity is enough for only two items. The objective coefficients (selling prices of each item) of the items are not known but we know the features $\vparam^1 = [-1,3]$, $\vparam^2 = [0,1]$, $\vparam^3 = [1,1]$. 
We have a linear model to predict selling prices using the given features. Its parameters are $\vcoef = [\coef_{1},\coef_{2}]$ and $\vpredsmall ^i = \coef_{1} * \param_{1}^i + \coef_{2} * \param_{2}^i + c$ where $i\in\mathbb{Z}, 0<i<4$. Assume $\coef_{2}$ is fixed and is equal to 1, and for simplicity the constant $c$ is also equal to 0. Then we can express the predicted objective coefficients with linear functions (see Figure~\ref{fig:PWL}), $\vpredsmall_{1} = -\coef_{1} + 3$, $\vpredsmall_{2} = 1$, $\vpredsmall_{3} = \coef_{3} + 1$. 
If we inspect Figures~\ref{fig:PWL} and \ref{fig:PWLsum} we can see that although $\vpredsmall$ continuously changes with $\coef_{1}$, there are only three different solutions ($\xopt \in \{(1,1,0), (1,0,1), (0,1,1)\}$)
provided by the solver. Each different solution represents a sum of linear functions of the chosen items. By combining all the separate linear functions, we represent the complete solution space as a piece-wise-linear function (PWLF) seen in Figure~\ref{fig:PWLsum}.}
\end{exmp}

 In example \ref{ex_knap} there are two transition points; $\coef_1=0$, $\coef_1=2$.  We can express the solution space of this 0-1 knapsack problem as
\[ \begin{cases} 
      \vpredsmall_1,\vpredsmall_2\geq \vpredsmall_3,  s(\vparam,\coef_1) \equiv   (1,1,0) & \coef_1\leq 0 \\
      \vpredsmall_1,\vpredsmall_3\geq \vpredsmall_2,  s(\vparam,\coef_1) \equiv   (1,0,1) & 0\leq \coef_1\leq 2 \\
      \vpredsmall_2,\vpredsmall_3\geq \vpredsmall_1,  s(\vparam,\coef_1) \equiv   (0,1,1) & 2\leq \coef_1 
   \end{cases}
\]

We define the \emph{predicted optimal value} (POV) and \emph{true optimal value} (TOV), for fixed feature value $\vparam$, as follows: 

\begin{align*}
     POV(\vcoef) = Obj(s(\vparam,\vcoef), \vpred(\vparam, \vcoef)) \\ 
     TOV(\vcoef) = Obj(s(\vparam,\vcoef),\vopt)  
\end{align*}

Note that the predicted coefficients do not directly affect the true objective value shown in Figure~\ref{fig:PWLopt}. 
However the transition points of the predicted PWLF are 
exactly the same as the transition points of the discrete, true objective. Therefore, if we identify transition points of the predicted function, we can use them to dramatically reduce the effort to map the real objective function.


Our framework works for any arbitrary optimization problem with a linear objective function. An optimization problem has a linear objective function when the relationship between the solution vector and the coefficients are linear:
\begin{equation}\label{eq_linear_objective}
    Obj(\xopt,\vopt) = \xopt^T \cdot \vopt
\end{equation}
We also assume predicted coefficients are parameterised linear functions
\begin{equation}\label{eq_parameterized_linfunc}
    \vpredsmall = \vcoef^T\cdot\vparam + c
\end{equation}
\noindent 
where $c$ is a constant.
This is the same restriction as~\citet{aaai20e}.
Equations \ref{eq_linear_objective} and \ref{eq_parameterized_linfunc} mean that the predicted objective value of the optimization problem can be expressed as a sum of linear functions. 

We now discuss two properties of $POV$ assuming $\vcoef$ is a singleton $\beta$.
Since we shall use coordinate descent to reason about $POV$ while modifying only one parameter, this is the only case we are interested in.

\begin{lemma}\label{lemma_pwlf}
     $POV$ is a piecewise linear function.
    \begin{proof}
    $POV(\beta) = s(\vparam,\beta)^T \cdot \vpred(\vparam,\beta) 
    = (\argmax_{\xopt \in C} \vpred(\vparam,\beta) \xopt)^T \cdot \vpred(\vparam,\beta) 
    = \max_{\xopt \in C} \xopt^T \cdot \vpred(\vparam,\beta)$
    Since it is the max of a set of linear functions, it is piecewise linear.
    \end{proof}
 \end{lemma}

\begin{lemma}\label{lemma_der}
     $POV$ is a convex function.
    \begin{proof}
       We need to show that $POV(t \beta_1 + (1-t) \beta2) \leq t POV(\beta_1) + (1-t) POV(\beta_2)$ for $t \in [0,1]$, $\beta_1,\beta_2$ arbitrary.
       Define $\beta_3 = t \beta_1 + (1-t)\beta_2$. For $i \in \{1,2,3\}$, let
       $\xopt_i = s(\vparam,\beta_i)$, so $\xopt_1^T \cdot \vpred(\vparam,\beta_1) \geq \xopt_3^T \cdot \vpred(\vparam,\beta_1)$ as $\xopt_1$ is the arg max, and similarly 
       $\xopt_2^T\cdot \vpred(\vparam,\beta_2) \geq \xopt_3^T \cdot \vpred(\vparam,\beta_2)$.
       Now
       \begin{eqnarray*}
       && POV(t \beta_1 + (1-t) \beta_2) \\
       &= & \xopt_3^T \cdot \vpred(\vparam, t \beta_1 + (1-t) \beta_2)  \\
       & = & \xopt_3^T \cdot ((t \beta_1 + (1-t) \beta_2)\vparam + c)  \\
       & = & \xopt_3^T \cdot (t (\beta_1 \vparam + c) + (1 - t)(\beta_2 \vparam + c)  \\
       & = & t \xopt_3^T \cdot \vpred(\vparam,\beta_1)  + (1-t) \xopt_3^T \cdot  \vpred(\vparam,\beta_2)  \\
       & \leq & t \xopt_1^T \cdot \vpred(\vparam, \beta_1) + (1-t) \xopt_2^T \cdot  \vpred(\vparam, \beta_2)  \\
       & = &  t POV(\beta_1) + (1-t) POV(\beta_2)
       \end{eqnarray*}
    \end{proof}
 \end{lemma}

\begin{corollary}\label{corr_line}
    For any three values $\coef_1<\coef_2<\coef_3$, 
    the points $(\beta_1,POV(\beta_1)), (\beta_2, POV(\beta_2)), (\beta_3, POV(\beta_3))$ are not collinear iff there is a transition point $\beta_t$
    in the range $\coef_1<\coef_t<\coef_3$. 
    \begin{proof}
        Since $POV$ is piecewise linear (Lemma~\ref{lemma_pwlf}), if three points are not collinear, there must be a transition point between them.
        Given $POV$ is convex (Lemma~\ref{lemma_der}) if three points are collinear there can be no transition point between them.
        \end{proof}
\end{corollary}

   
  


\subsection{Transition point extraction}\label{extraction}


\citeay{aaai20e} represent optimization coefficients as parameterised linear functions. They solve the optimization problem with dynamic programming using piece-wise linear algebra and parameterised coefficients. The output of the DP solution gives the piecewise linear function POV. However this method is limited to problems with a DP solution. In addition, the DP solution requires constructing the full PWLF for every problem set. For larger problems this may result in long run times. Instead we propose a numerical approach to extract transition points. Our approach works for any arbitrary optimization problem with a linear objective. 
\ignore{We also propose greedy methods that partially construct the PWLF, and show that partial mappings can decrease run time and still achieve the same performance as the full methods.
}



\ignore{
\aug{-PWLF is continuous}

\pjs{Dont follow this proof}
\begin{lemma}\label{lemma_cont}
        POV is a continuous function $\forall\coef$.
        \pjs{Continuous means continuous at all points}
        \begin{proof}
            Let $\coef_t$ be a transition point, $\forall$ $\epsilon>0$ we have\\  $POV(\vpred(\coef_t-\epsilon),s(\coef_t-\epsilon))>POV(\vpred(\coef_t-\epsilon),s(\coef_t+\epsilon))$\\ $POV(\vpred(\coef_t+\epsilon),s(\coef_t+\epsilon))>POV(\vpred(\coef_t+\epsilon),s(\coef_t-\epsilon))$\\
            when we fix the solutions and let $\lim_{\epsilon\to0^-}$ and $\lim_{\epsilon\to0^+}$
            $POV^-(\vpred(\coef_t),s(\coef_t-\epsilon))>POV^-(\vpred(\coef_t),s(\coef_t+\epsilon))$\\ 
            $POV^+(\vpred(\coef_t),s(\coef_t+\epsilon))>POV^+(\vpred(\coef_t),s(\coef_t-\epsilon))$\\
            POV is a sum of linear functions, then at $\coef=\coef_t$\\
            $POV(\vpred(\coef_t),s(\coef_t+\epsilon))=POV(\vpred(\coef_t),s(\coef_t-\epsilon))$\\
        \end{proof}
        
\end{lemma}
}

\begin{figure}
    
    \centering
    \caption{Divide and conquer algorithm}
    \begin{subfigure}[t]{.13\textwidth}
        \centering
         \includegraphics[width=0.8\textwidth]{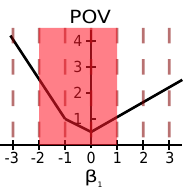}
        \caption{Sample points with large and uniform step size. The algorithm finds a transition point in the interval [-2,1]}
        \label{fig:dnc1}
    \end{subfigure}%
    \hspace{1em}
    \begin{subfigure}[t]{.13\textwidth}
        \centering
        \includegraphics[width=0.8\textwidth]{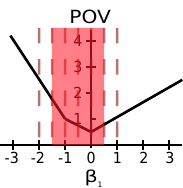}
        \caption{The algorithm decreases the step size and narrows the transition point to the interval [-1.5,0.5]}
        \label{fig:dnc2}
    \end{subfigure}
    \hspace{1em}
    \begin{subfigure}[t]{.13\textwidth}
        \centering
        \includegraphics[width=0.8\textwidth]{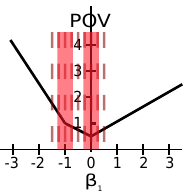}
        \caption{With a precise step size, the algorithm identifies two transition points in the intervals [-1.25,-0.75] and [-0.25,0.25]}
        \label{fig:dnc3}
    \end{subfigure}
    \label{fig:dnc_all}
\end{figure}

\textbf{Divide and conquer:}
From corollary \ref{corr_line}, we know that if we compare arbitrary three points of POV and they are not collinear, then there is at least one transition point between those values. When we decrease the distance between these three values, we can accurately identify the location of transition points. The simplest way to extract transition points is to sample POV with a fixed step size, and compare collinearity of consecutive three points. Clearly this brute force method can be infeasible for many problems except the easiest ones. Especially for long intervals without transition points, using a small step size is redundant. With these insights in mind we apply a divide and conquer algorithm to sample POV. First we split the search region with ten uniformly spread points, then we test collinearity of these points. If we find any points that are not collinear, we mark the intervals defined by these points as \textbf{transition intervals}. Finding a transition interval means that there is at least one transition point in the interval. Then we proceed to decrease the step size as: $step_{n+1} = \frac{step_n}{10}$ and sample the transition intervals. Finally we repeat finding transition intervals, and reducing the step size until the step size reaches a desired minimum. By starting with a large step size and iteratively reducing it, we identify long intervals without transition points with minimal processing (Fig ~\ref{fig:dnc_all}).


\textbf{Coordinate Descent:} We described a method to construct the piece-wise linear function POV and extract transition points by comparing collinearity of sample points. However we assumed single dimensional parameterised coefficients. In reality multi-variate models are widely used to predict these coefficients. We use \textbf{coordinate descent} to transform a multivariate linear model into a  one dimensional model \cite{wright2015coordinate}. For a parameter vector $\vcoef = [\coef_1,...,\coef_m]$, coordinate descent iterates over $\vcoef$. In each iteration one parameter, $\coef_k$, is considered as a variable, and the rest are fixed as a constant, $\vpredsmall=\coef_k\cdot\param_k + \sum_{n\neq k}\coef_n\cdot\param_n$. Then for each parameter we consecutively perform transition point extraction and parameter updates.

After all transition points are identified we proceed to pick the best overall transition intervals to update model parameters. This process is explained in detail in the next section.



\subsection{Parameter Update}\label{param_update}
In a predict+optimize setting a dataset is a collection of multiple problem sets. Each problem set has the same constraints for the optimization problem, however their coefficients are different. We predict unknown coefficients for each problem set with the same model parameters, and the goal of the framework is to train model parameters to minimize the average regret across all problem sets. For a dataset of size $N$, and coefficient vectors $\vopt$, the dataset is denoted $D=\{\vopt^{(1)},...,\vopt^{N}\}$. We choose the model parameters $\vcoef$ to minimize the average regret $R$.
$$\vcoef \equiv \argmin\frac{1}{N}\sum_{n=0}^{N} R(\vopt^{(n)},\vpred^{(n)}) $$



\textbf{Transition point comparison:} 
We express the true objective value (TOV) of each problem set as a piece-wise function. The extraction method provides the transition points of each problem set. Let $T^{(i)}=\{\coef^{(i)}_{t_1},...\coef^{(i)}_{t_L}\}$ be the set of transition points of size $L$ for a problem set $i$. We construct the intervals of the piece wise function as $I^{(i)}=\{[\coef^{(i)}_{t_l},\coef^{(i)}_{t_(l+1)}],0<l<L,l\in\mathbb{Z}\}$. To find the optimal model parameters for a problem set $i$, we can simply calculate TOV for each interval and pick the best interval. As a single sample point for each interval is enough to calculate TOV, we choose the mid points of the intervals for calculations. Let $I_{mid}$ represent the set of mid points then
$\coef^{(i)}_{opt}=\argmin_{\coef\in I^{(i)}_{mid}} R(\vopt^{(i)},\vpred^{(i)})$. However, the optimal parameter for each problem set can be different. To find the optimal parameters over all problem sets, we compare over every interval from each problem set.
$$\coef_{opt}\equiv\argmin_{\coef\in\bigcup\limits_{i=1}^{N} I^{(i)}_{mid}
} \frac{1}{N}\sum_{n=0}^{N} R(\vopt^{(n)},\vpred^{(n)})$$
With coordinate descent we perform these comparisons for each parameter and update the parameters individually.
\textbf{Mini-batch: } In our framework we use mini-batches to train the model parameters. A mini-batch represents a subset of problem sets. When using mini-batches we construct a quasi-gradient in the direction of the global minimum of that particular mini-batch. Then we update the model parameters with the quasi gradient. For a mini-batch $i$:
$\coef_{new}=\coef_{old} + learning\_rate\cdot(\coef^{(i)}_{opt} - \coef_{old})$

\subsection{Greedy Methods}
To extract the transition points and compare them, we have to repeatedly solve an optimization problem. Many optimization problems are NP-hard, and obviously as the problem size increases it can become expensive to perform these actions. Therefore we propose a greedy method to partially extract transition points and a greedy method to compare only the best transition points.


\textbf{Divide and Learn Max (DnL-MAX) :}
Normally we compare all intervals for all problem sets. If we assume there are $L$ intervals for each problem set $i$ and a total of $N$ problem sets, then we solve the optimization problem $N^2L$ times. For a full method the complexity scales both with the number of transition points and the size of the data. Instead we propose to first choose the optimal parameter for each problem set $i$ by $\coef^{(i)}_{opt}=\argmin_{\coef\in I^{(i)}_{mid}} R(\vopt^{(i)},\vpred^{(i)})$, and then compare only the best parameters with the other problems sets:
$$\coef_{opt}=\argmin_{\coef\in\bigcup\limits_{i=1}^{N} \coef^{(i)}_{opt} }
\frac{1}{N}\sum_{n=0}^{N} R(\vopt^{(n)},\vpred^{(n)})$$ 
This reduces complexity by solving only $(N-1)N+LN$ optimization problems, and the number of transition points has minimal effect on the comparison complexity.


\textbf{Divide and Learn Greedy (DnL-Greedy) : } 
Our divide and conquer algorithm repeatedly compares the collinearity of POV samples. Depending on the complexity of the problem and the minimum step size for sample points, there can be many redundant transition points. We propose a greedy extraction method to stop the extraction at the first transition point $\coef_t$ that improves TOV over the old parameter $\coef_{old}$. We also prioritize regions around the old model parameter for transition point search. We observed that although TOV is not a convex function, the optimal model parameters are clustered in similar regions. Our motivation with this greedy method is to quickly iterate over parameters and bypass redundant sampling. Note that as we use only one transition point for each problem set, DnL-Greedy automatically includes the greedy comparison of DnL-MAX.

The greedy approaches do not guarantee  finding the global minimum. However we show empirically they achieve the same performance as the full method, 
and reduce the run time dramatically.
\section{Evaluation}

In this section we detail our experiments. We experiment on two optimization problems: 0-1 knapsack and scheduling. We run experiments for four exact methods: \emph{DnL}, \emph{DnL-MAX}, \emph{DnL-Greedy}, \emph{dynamic programming (DP)}~\cite{aaai20e}, two surrogate methods:
\emph{SPO}-Relax~\cite{aaai20a}, \emph{QPTL}~\cite{ferber2020mipaal} and one indirect method: \emph{ridge regression}.

\textbf{Dataset:} We use the dataset from the ICON energy challenge ~\cite{ICON} for both knapsack and scheduling problems.  Data samples are collected from real electricity prices every 30 minutes, from 2011 November to 2013 January. Wind forecast, wind speed, Co2 intensity, temperature, load forecast and price forecast are used to predict actual energy prices.
Note that predicting energy prices is challenging so even the best models have substantial prediction error. The same dataset was used in previous work on predict+optimize \cite{aaai20a,aaai20e}.

In total there are 37877 data samples, each 48 data samples, representing a day, create a problem set. Therefore we can only use 789 optimization problems to train predict+optimize models. We split the data set into 70\% train set, 10\% validation set and 20\% test set. 
Correspondingly we have 552, 79 and 157 optimization problems for training, validation and testing, respectively. In order to understand how models work for different distributions, we split the dataset into 5 folds. For each fold we train every model 10 times and use the iteration with the least validation regret. We report the performance of models over the mean and standard deviation of all folds.


\begin{figure}
    \centering
    \includegraphics[width=0.48\textwidth]{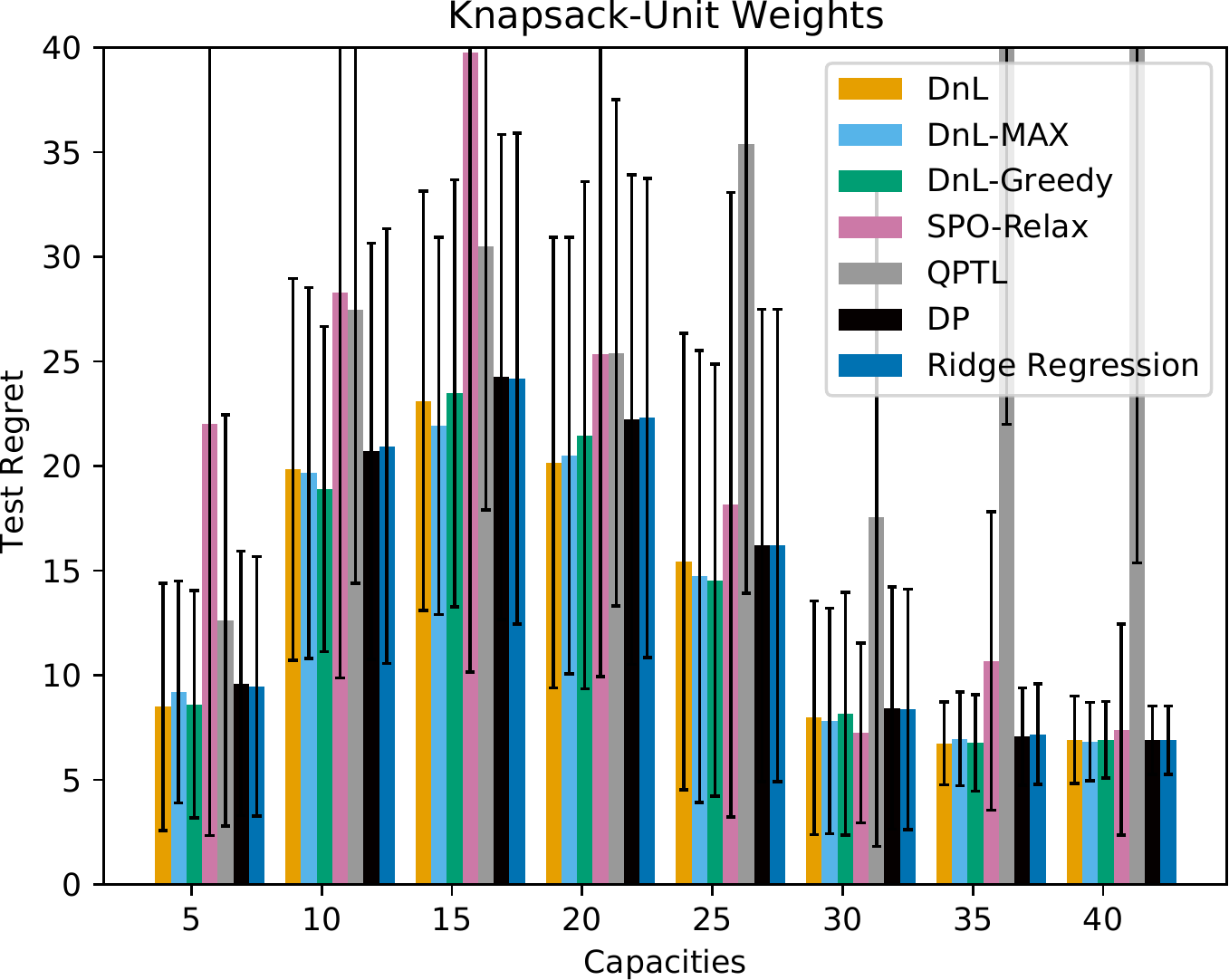}
    \caption{Unit Knapsack: showing average and one standard deviation. 
    Note that the graph is truncated at 40 for readability.}
    \label{fig:unit_knapsack_qptl}
\end{figure}

\begin{figure}
    \centering
    \includegraphics[width=0.48\textwidth]{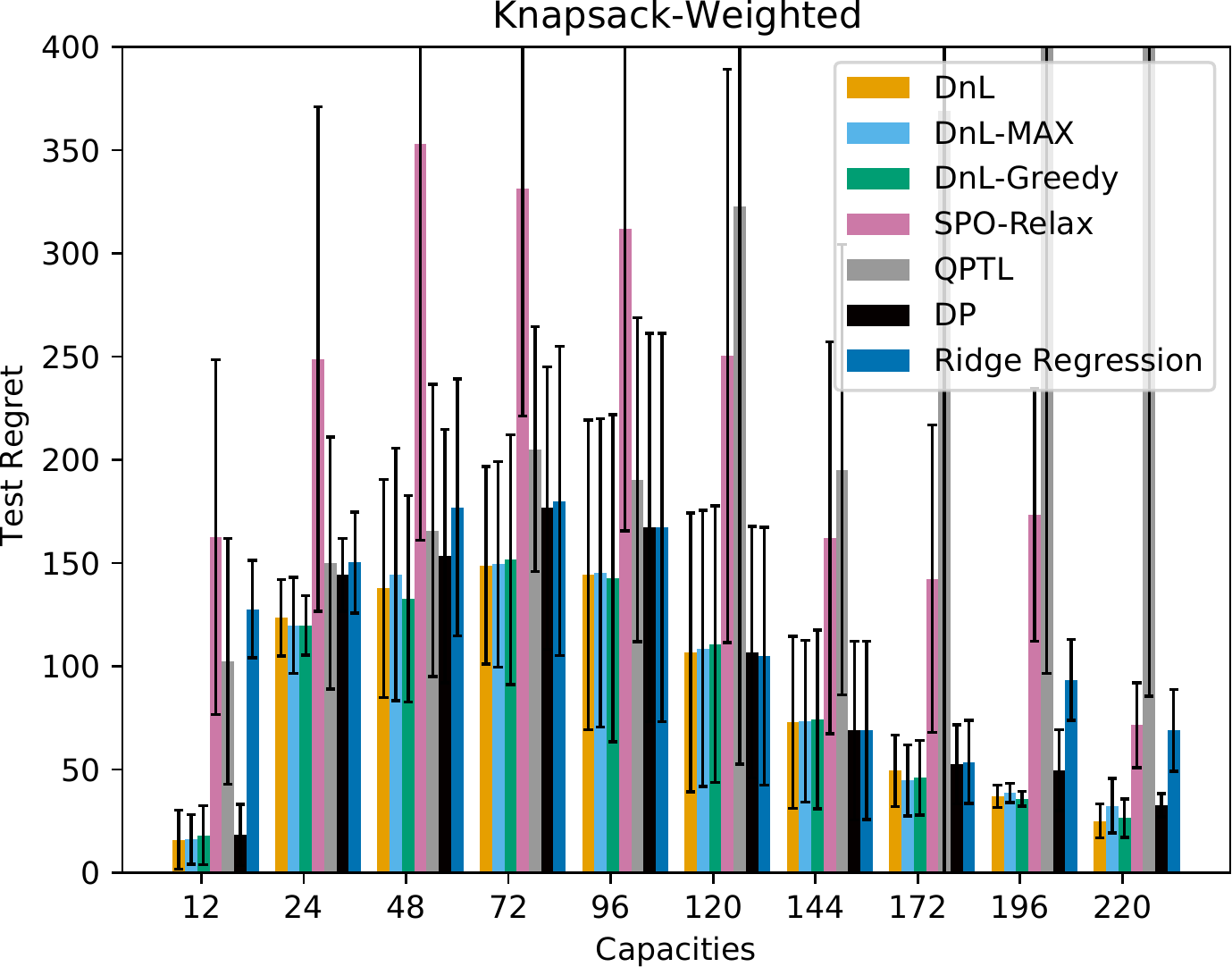}
    \caption{Weighted Knapsack: showing average and one standard deviation. SPO-Relax and QPTL are truncated at 400 for some instances for readability. }
    \label{fig:weighted_knapsack_qptl}
\end{figure}

\ignore{
\begin{figure}
    \centering
    \includegraphics[width=0.45\textwidth]{figs/unweighted_knapsack_qptlh.pdf}
    \caption{patterns}
    \label{fig:unit_knapsack_qptlh}
\end{figure}

\begin{figure}
    \centering
    \includegraphics[width=0.5\textwidth]{figs/strong_corr_w_knapsack_qptlc.pdf}
    \caption{colored error bars with patterns\aug{which one is better?}}
    \label{fig:unit_knapsack_qptlhc}
\end{figure}

\begin{figure}
    \centering
    \includegraphics[width=0.5\textwidth]{figs/strong_corr_w_knapsack_qptlcxh.pdf}
    \caption{colored error bars without patterns}
    \label{fig:unit_knapsack_qptlcxh}
\end{figure}
}

\textbf{Knapsack problem:} We consider a 0-1 knapsack of $n$ items, where we are given a capacity limit, $W$, item weight $\mathbf{w}=[w_1,w_2...,w_n]$, and have to predict
item values $\vopt=[v_1,v_2...,v_n]$. 
\ignore{
A 0-1 solution vector $\xopt$ represents the chosen items
$$\argmax_{\xopt^T \cdot \mathbf{w} \leq W } \xopt^T \cdot \vopt$$
}
A 0-1 solution vector $\xopt = \argmax_{\xopt^T \cdot \mathbf{w} \leq W } \xopt^T \cdot \vopt$ decides the chosen items. 
%
%
We run knapsack experiments for both unit weights and varying weights. The original dataset does not have item weights, and we generate the weights synthetically.
Knapsack problems with high correlation between item value and weight
are considerably harder to solve than those with weak correlation~\cite{pisinger2005hard}.
We create exactly correlated knapsack problems by choosing weight values in $\{3,5,7\}$,
and multiplying by the true energy price to generate their true value. We experiment on varying capacity limits from 5\% to 90\%. For unit knapsack the capacity limits range from 5-45 (10\%-90\%) and for weighted knapsack the capacity limits range from 12-220 (5\%-90\%).

\ignore{
\textbf{Weight generation: } A knapsack problem with high correlation between item values and weight, i.e. the profitability of items are close to 1, is called a \emph{strongly correlated} knapsack problem. Strongly correlated knapsack problems have a larger gap between their continuous and integer solutions. Therefore strongly correlated knapsack problems are considered harder to solve compared to weakly correlated ones~\cite{pisinger2005hard}. To emphasize our models' ability to capture harder combinatorial problems over surrogate models, we use an exactly correlated knapsack problem where all items have the same profitability ratio. To generate an exactly correlated knapsack we create a weight vector $\pmb{w}=[w_1,...,w_n]$ by choosing weight values from a seed of [3,5,7]. The weight vector has equal distributions of weights, and the total sum of weights is 244. Then we generate the correlated energy prices by multiplying the weight vector and the original energy prices $\vopt_{weighted} = \vopt_{unit}{\pmb{w}}$. We treat weights as an additional feature for predictions. We experiment on varying capacity limits from \%5 to \%90. For unit knapsack the capacity limits range from 5-45(\%10-\%90) and for weighted knapsack the capacity limits range from 12-220(\%5-\%90).
}

\textbf{Scheduling:}
We test on energy cost aware scheduling problems. The scheduling problems are a simplified versions of the ICON challenge~\cite{aaai20a}.
There are $M$ machines and $N$ jobs. Each machine has a resource capacity $C_m$. Each job has a resource requirement $R_n$, power consumption $P_n$ and a duration $D_n$. Every job also has an earliest starting time $te_n$ and a latest finishing time $tl_n$. A job can only be run on one machine, and once a job is being processed it cannot be split. All jobs have to be finished in the 24 hour period. The goal of the scheduling is to minimize energy costs by taking energy prices into account. Energy prices are not known beforehand and have to be predicted. We consider three different loads with 3,5,3 machines and 15,20,10 machines correspondingly. 


\ignore{
\textbf{Divide and learn (Dnl) :} We experiment on three versions of Dnl: DnL-FULL, is the full version of divide and learn framework, DnL-MAX uses only the best transition point of each problem set for comparison and DnL-Greedy is the model with the greedy transition point extraction. We set the minibatch size to 32 and learning rate to $0.1$. We initialize the model parameters with ridge regression. The search space for transition points are bounded relative to the each parameter as $[\coef-1.5\coef,\coef+1.5\coef]$. Similarly the minimum step size of models are relative to the parameter and is equal to $step_{min} = \coef/10$.
}

\subsection{Experiments}

The models are trained with Intel(R) Xeon(R) Gold 6254 CPU @ 3.10GHz processors using 8 cores with 3.10 Ghz clock speed. We use~\citeay{gurobi} to solve knapsack and scheduling problems. For the knapsack problems max training time is set to 4000 seconds ($\approx$ 1 hour). For scheduling problems max training time is set to 12000 seconds ($\approx$ 3.3 hours). We tune hyper-parameters via grid search for surrogate models and we use early stopping for all models ~\cite{bishop2006pattern}. DnL hyper-parameters are as follows:  We set the mini-batch size to 32 and learning rate to $0.1$. We warmstart the model parameters with ridge regression \cite{pratt1996survey}. The search space for transition points are bounded relative to the current value of the parameter as $[\coef-1.5\coef,\coef+1.5\coef]$. Similarly the minimum step size of models are relative to the parameter and is equal to $step_{min} = \coef/10$.

\textbf{Unit knapsack: } Unit knapsack (Figure~\ref{fig:unit_knapsack_qptl}) is a relatively simple optimization problem and there is not a large difference between regression and predict+optimize models. All variations of DnL perform identically to DP, and this suggests greedy approaches perform as well as their full counterparts. Unlike regression and exact methods, the surrogates' performances are sensitive to the changes. SPO-Relax fails to capture the optimization problem for low capacities, while QPTL fails to capture the high capacity problem.

\begin{figure}
    \caption{Low capacity knapsack, test regret vs epoch plots,
    }
    \centering
     \begin{subfigure}[b]{.22\textwidth}
        \centering
        \includegraphics[width=1\textwidth]{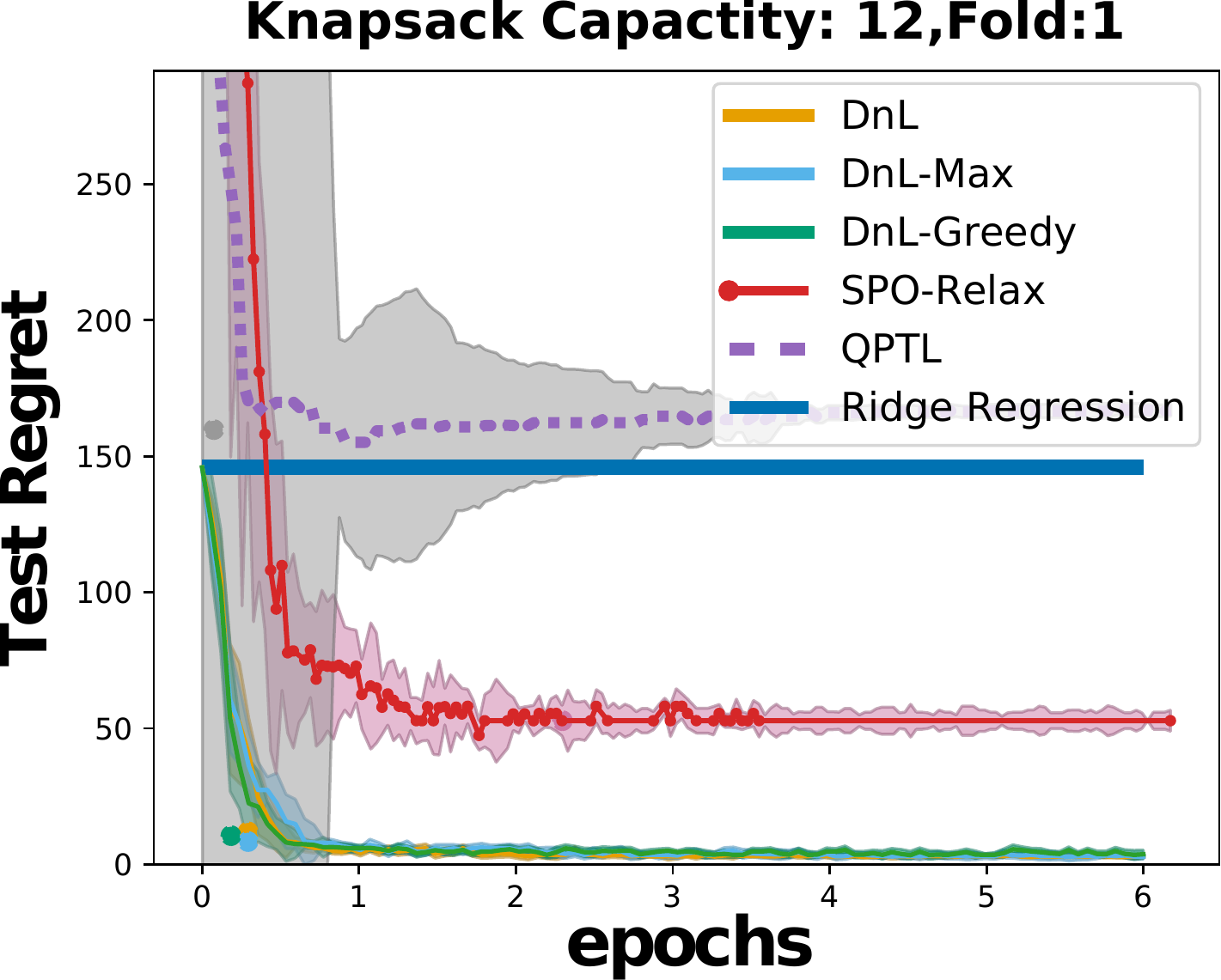}
        \caption{weighted knapsack capacity 12,fold 1. SPO-Relax converges to low regret, QPTL overfits}
        \label{fig:kc12k1.pdf}
    \end{subfigure}%
     \hspace{1em}
    \begin{subfigure}[b]{.22\textwidth}
        \centering
        \includegraphics[width=1\textwidth]{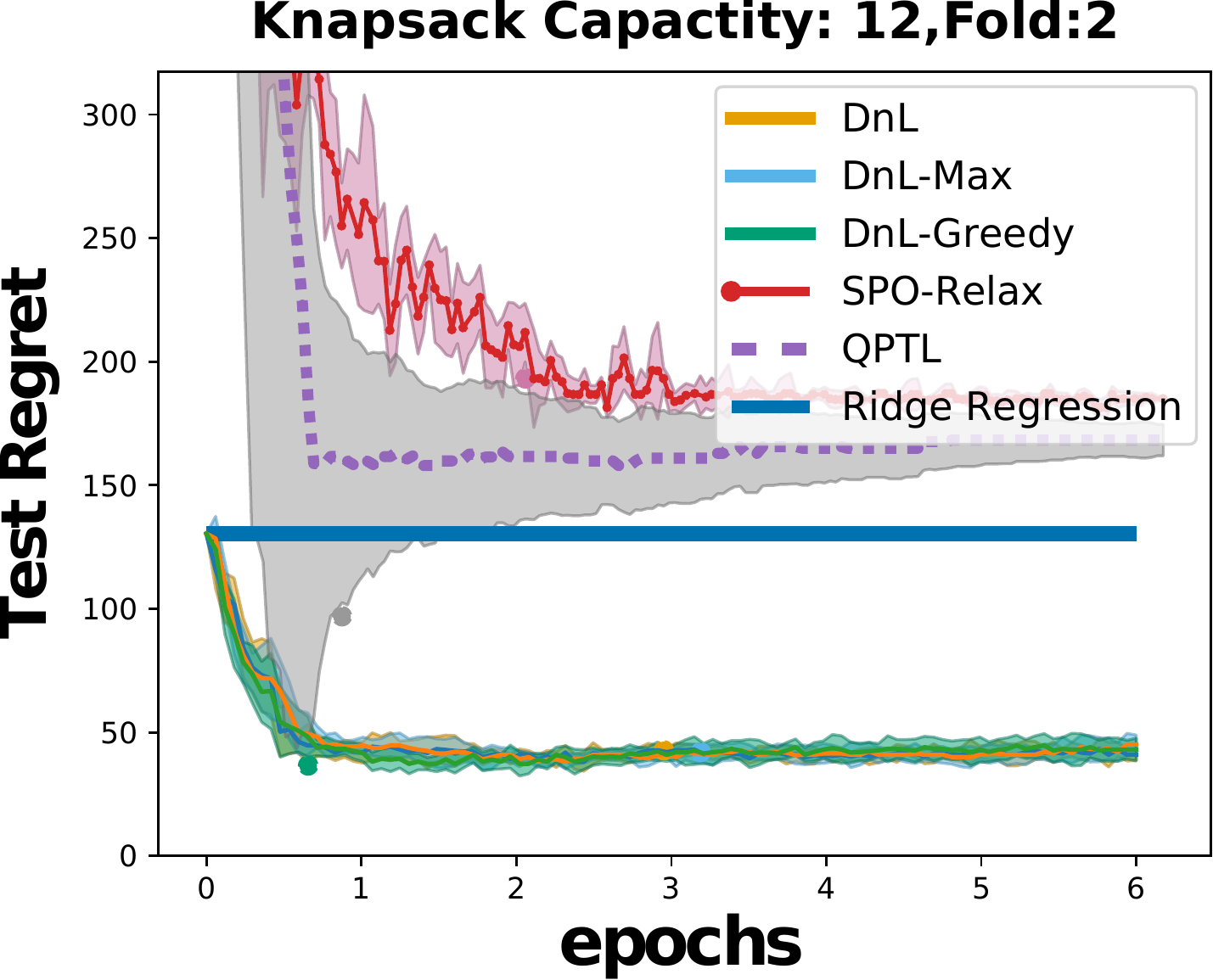}
        \caption{weighted knapsack capacity 12,fold 2. SPO-Relax converges to high regret, QPTL overfits}
        \label{fig:kc12k2.pdf}
    \end{subfigure}
    \label{fig:c12converge}
\end{figure}

\textbf{Weighted knapsack: } 
The exact methods outperform surrogates for all capacities and regression for low and high capacities (Figure~\ref{fig:weighted_knapsack_qptl}). 
Their advantage is more clear at the extreme capacities \{12,196,220\}. 
For capacities \{72,96,196\} DnL outperforms DP. At these capacities DP requires long time to build the piece-wise function and fails to finish an epoch within the time limit. In contrast for the same weights DnL-Greedy can train an epoch under 10 seconds. Similar to the unit knapsack problem SPO-Relax has higher regret for low capacities. We have observed that SPO-Relax is very sensitive to the data distribution for the weighted knapsack problem. For some folds it successfully converges at a low regret minimum. However for other folds, it converges at a higher regret than regression (Fig ~\ref{fig:c12converge}). QPTL outperforms regression for low capacities but for high capacities it has the worst regret of all models. QPTL also tends to overfit the problem, though early stopping helps to identify a minimum (Fig~\ref{fig:c12converge}). These experiments show that exact methods are able to accurately search the underlying knapsack problem for all capacities, and they are more robust to changes in constraints.



\begin{table}[h!]
\small
\centering
\resizebox{\columnwidth}{!}{%
\begin{tabular}{||l c c c ||} 
 \hline
 Loads & 1 & 2 & 3 \\ [0.5ex] 
 \hline\hline
 DnL & $8382\pm{3616}$ & $17168\pm{7292}$ & $\pmb{10347\pm{5321}}$  \\ 
 DnL-Greedy & $8625\pm{3829}$ & $15834\pm{6259}$ & $10895\pm{5626}$\\
 SPO-Relax & $8421\pm{4144}$ & $18454\pm{8302}$ & $11086\pm{4752}$ \\
 QPTL& $6584\pm{2157}$ &$27393\pm{12606}$ &$10544\pm{5138}$\\
 Regression&$\pmb{6544\pm{4360}}$&$\pmb{15672\pm{6542}}$&$10919\pm{6086}$\\
 [1ex] 
 \hline
\end{tabular}
}
\caption{ICON scheduling problems mean regret and standard deviations}
\label{table:1}
\end{table}

\textbf{Scheduling: } 
Compared to knapsack problems scheduling problems are more complex and do not have a dynamic programming solution. As such DP method is not applicable for these problem sets and to the best of our knowledge DnL is the first exact method applicable for an arbitrary MIP problem with a linear objective.
Although all frameworks have different loss models, they all converge at a similar point and we do not identify one method that clearly outperforms all the others. We believe the underlying problem for these loads, like medium capacity knapsack problems, are similar to MSE and  understanding the underlying regression problem is also a successful result for predict+optimise frameworks.
For Load 1, regression has the best regret, QPTL is the best predict+optimize framework, and the other three have similar regrets. Interestingly for some folds of Load 2, QPTL cannot reason over regret. We believe the solution space of these folds may resemble the high capacity knapsacks. We also observe that SPO-Relax has a higher regret than regression for Load 2. Although DnL seems to have a high regret as well, this is a result
of slow training and not completing within the 4 hour limit. The greedy variation is able to complete the training and has a similar regret to the regression. For Load 3, we do not see a significant difference between the frameworks. 

\ignore{
\subsubsection{Underlying Problem}
As the underlying problem diverges from a MSE estimation, the value of using predict+optimize frameworks increases. For example if we observe the weighted knapsack problem, intuitively we can say that for low capacities the underlying problem is to predict high valued items. For average capacities, to predict the average valued item and for high capacities to predict low valued items. And if we check figures ~\ref{fig:weighted_knapsack_qptl} we can see that predict+optimize frameworks are improving over regression for low and high capacities. We believe the underlying problem of scheduling problems are similar to regression. We have experimented with different variations of machine capacities and job requirements, however we couldn't find a meaningful relationship. Note that for combinatorial problems small changes in constraints can change the solution space drastically and for complex problems without running predict+optimize frameworks it maybe hard to understand if the underlying problem is vastly different than regression. However from our experiments we think exact methods are able to improve over regression when the underlying problem is different, and at worse they are able to capture the regression problem without too much difference.

\subsection{Fit to surrogate}
Even when the underlying problem is different than a standart ML loss, surrogate models have to be careful about how this problem fits the surrogate. From knapsack experiments we see that two different surrogates have opposite behaviours for the same problem. While QPTL has good performance for low capacities, it fails for high capacities. SPO-Relax cannot reliably understand low capacities but their performance increases for high capacities. Similarly for electric scheduling the performances of surrogates can change vastly when different loads are used. While QPTL is the best framework for load1, we have seen it diverges for load2. In contrast exact problems do not change their behaviour as drastically with the constraint changes. Therefore it is better to use surrogates for well known problems, and if that surrogates has been shown to fit that particular problem. As we stated combinatorial problems can have very different solution spaces with small constraint changes, and exact methods are likely to be more robust against such unknown changes. 
}

\subsubsection{Scalability/runtime}
Scalability is the biggest issue for exact methods. Combinatorial optimization problems are expensive to solve, and exact methods require solving multiple instances to map each problem set. Therefore we proposed greedy methods, and observed improved efficiency over the existing exact method DP. For example, DP requires more than an hour to train an epoch for knapsack capacities \{72,96,196\} and DnL-Greedy can complete the training of an epoch under 10 seconds. For other capacities, DP requires 200--400 seconds for an epoch, while we require 2--10 seconds. There are two reasons for this. First, the DP method has to make use of a dynamic 
programming solution compatible with piece wise linear algebra, which may not be the 
most efficient solution approach. Second, the DP constructs the entire piecewise function every time. Since the number of transition points can increase exponentially with problem size, our greedy methods can avoid computing irrelevant transition points.

Although DnL-Greedy is more scalable than the previous exact methods, it requires more processing power than surrogate methods. This difference is emphasized for larger problems. For scheduling problems (load3) DnL-Greedy requires 20 and DnL requires 30--60 minutes to train an epoch. By comparison SPO-Relax, which uses the relaxation of the optimization problem, can finish an epoch in under a minute. 
For a fair comparison we ran SPO-Relax for the same amount of time that we ran DnL-Greedy, but we did not observe any change in the convergence behaviour or  the output regret.

\ignore{

\aug{Maybe a comparison table might be good?}

\subsubsection{Embarassingly Parallel}
\aug{I can talk about the parallelization of Dnl here. Or we can ignore it.}

\subsection{How to approach predict+optimize}
 \aug{I think I will delete this section, or change it to future discussion\ or conclusion}
From our experiments we think not all predict+optimize problems are the same. Characterizing combinatorial optimization problems can be quite hard, small changes in constraint and problem sets can result in completely different behaviour. Here we try to provide possible guidelines  on different types of predict+optimize problems
\subsubsection{Underlying Problem}\label{regvsreg}
First step to characterize a predict+optimize problem is to understand the underlying problem. When the MSE predictions are accurate enough, or when the problem space is not very combinatorial regression models can have similar performances with specialized predict+optimize frameworks. For such problems it is better to use simpler and cheaper regression models. However the challenge is how to identify the problems before training predict+optimize models. Although specific well-known problems can be easier (knapsack) to characterize, to the best of our knowledge there is no methodological way to understand when predict+optimize outperforms MSE.


\subsubsection{Combinatorialness}
Second step after deciding whether to use predict+optimize is to see if a surrogate fits the combinatorial problem. If we know surrogates would fit adequately to the combinatorial problem, we do not have to use expensive exact methods. However similar to the underlying problem, to the best of our understanding there is no methodological to decide which problems fit surrogates and which do not.
\subsection{Algorithm Selection}
Choosing a predict+optimize framework comes with it's own costs and risks. Exact methods are expensive, however they have the best ability to understand predict+optimize problems. When an application deals with an unknown problem, and the performance is critical it might be worth-wile to invest in exact methods. If the optimization problem is studied well, and the problem is known to fit surrogates or the underlying problem is similar to MSE then we might choose to use cheaper indirect methods, or surrogates. When the dataset is large enough and MSE predictions are accurate, there might not be a need for predict+optimize too. 
}


\section{Conclusion}
Predict+optimize problems are challenging due to the combinatorial nature of the optimization problem.  We propose a new method to train parameters using regret, rather than a surrogate or relaxation. 
The only previous existing method requires a DP formulation of the optimization problem,
in contrast ours can be applied to any optimization problem with a linear objective.
By using greedy methods to find transition points we can substantially reduce the amount of search required. Our framework outperforms regression and surrogates for weighted knapsack problems and unlike the previous DP method, is able to reason over more complex problems.

\bibliography{reference.bib}

\end{document}